# Left-Center-Right Separated Neural Network for Aspect-based Sentiment Analysis with Rotatory Attention

**Shiliang Zheng, Rui Xia**[*]
School of Computer Science and Engineering,
Nanjing University of Science and Technology
zhengshiliang0@gmail.com, rxia@njust.edu.cn

## Abstract

Deep learning techniques have achieved success in aspect-based sentiment analysis in recent years. However, there are two important issues that still remain to be further studied, i.e., 1) how to efficiently represent the target especially when the target contains multiple words; 2) how to utilize the interaction between target and left/right contexts to capture the most important words in them. In this paper, we propose an approach, called left-center-right separated neural network with rotatory attention (LCR-Rot), to better address the two problems. Our approach has two characteristics: 1) it has three separated LSTMs, i.e., left, center and right LSTMs, corresponding to three parts of a review (left context, target phrase and right context); 2) it has a rotatory attention mechanism which models the relation between target and left/right contexts. The target2context attention is used to capture the most indicative sentiment words in left/right contexts. Subsequently, the context2target attention is used to capture the most important word in the target. This leads to a two-side representation of the target: left-aware target and right-aware target. We compare our approach on three benchmark datasets with ten related methods proposed recently. The results show that our approach significantly outperforms the state-of-the-art techniques.

## 1 Introduction

Aspect-based sentiment analysis is a fine-grained classification task in sentiment analysis, identifying sentiment polarity of a sentence expressed toward a target [Pang and Lee, 2008; Liu, 2012; Pontiki *et al.*, 2014]. In the early studies, methods for the aspect-based sentiment classification task were similar as that used in standard sentiment classification task. Researchers normally designed a set of features (such as bag-of-words, sentiment lexicons, and linguistic features) to train a statistical learning algorithm for sentiment classification [Kiritchenko *et al.*, 2014; Wagner *et al.*, 2014;

Vo and Zhang, 2015]. However, such kind of feature engineering work was labor-intensive and almost reached its performance bottleneck. In recently years, more and more researchers have adopted more advanced deep learning algorithms. By taking advantage of the powerful representation ability, well-designed neural networks can automatically generate meaningful low-dimensional representations for the targets and their contexts, and obtained the state-of-the-art results in aspect-based sentiment classification task [Dong *et al.*, 2014; Wang *et al.*, 2016; Tang *et al.*, 2016a; 2016b].

As we have mentioned, aspect-based sentiment classification differs from traditional sentiment classification in that the former is target-related. Jiang et al. [2011] pointed out that 40% of the classification errors are caused by ignoring the target information in twitter sentiment classification. The sentiment polarity of a sentence is strongly related to its target in aspect-based sentiment analysis. Taking the following sentence

Example 1: *"I am pleased with the life of battery, but the windows 8 operating system is so bad."*

for example, the target set is {*the life of battery*, *windows 8 operating system*}. As far the target *the life of battery* is considered, the expected sentiment is positive; by contrast, as far as the target *windows 8 operating system* is considered, the correct sentiment should be negative. That is, in one review sentence, the sentiment toward different targets could be opposite. Along with the deepening of research work, incorporating the target information into the model gradually becomes a consensus in aspect-based sentiment classification in recent years.

However, the previous way of modeling targets and contexts still have some shortcomings. For one thing, according to our statistics, more than 25% of the target on the Restaurant and 35% of the target on the Laptop datasets contain at least two words, but almost all researchers ignore the case of target phrase that contains multiple words, and just used the average of target constituting word vectors to represent target. For instance, Tang et al. [2016a] proposed a target-connection long short-term memory (LSTM) model, which utilizes the connections between target and each context word when composing the representation of a sentence. For another, the rep-resentations of targets and contexts are influenced by each other which is paid not enough attention. Taking Example 1 for example, with respect to the target "*the life of battery*", "*pleased*" should be paid with more attention than the other targets not related words (such as "*bad*") in the context; as for targets, "*life*" and "*battery*" should pay more attention in the representation of target "*the life of battery*". We can see that the representations of contexts are related to targets, meanwhile it is natural that targets are influenced by their contexts.

---
[*]The corresponding author of this paper.

In summary, when employing deep neural networks for aspect-based sentiment classification, the following two problems remain to be further studied:

- Problem 1: how to more efficiently represent the target especially when the target contains multiple words?
- Problem 2: how to utilize the interaction between targets and contexts to capture the most important words in the representation of targets and contexts?

With the attempt to better address the two problems, in this paper we propose a left-center-right separated neural network with rotatory attention mechanism (LCR-Rot). Specifically, we design a left-center-right separated LSTMs that contains three LSTMs, i.e., left-, center- and right- LSTM, respectively modeling the three parts of a review (left context, target phrase and right context). On this basis, we further propose a rotatory attention mechanism to take into account the interaction between targets and contexts to better represent targets and contexts. The target2context attention is used to capture the most indicative sentiment words in left/right contexts. Subsequently, the context2target attention is used to capture the most important word in the target. This leads to a two-side representation of the target: left-aware target and right-aware target. Finally, we concatenate the component representations as the final representation of the sentence and feed it into a softmax layer to predict the sentiment polarity.

The key characteristics of our work can be summarized as follows:

1. With respect to Problem 1, the target phrase is modeled with two-side representation which is combination of left-aware target and right-aware target. It better support the multi-word targets and leads to a significant improvement of classification performance;

2. With respect to Problem 2, the rotatory attention mechanism could utilize the interaction between targets and contexts to better represent targets and contexts;

3. We achieve currently the best aspect-based sentiment classification performance on three benchmark datasets. And we will release our code soon.

## 2 Related Work

Aspect-based sentiment analysis is a fine-grained classification task in sentiment analysis, which aims at identifying the sentiment polarity of a sentence expressed towards a target. In this work, we focus on the aspect term polarity detection task defined in SemEval 2014: for a given set of labeled aspect terms within a sentence, determine the polarity of each aspect term [Pontiki *et al.*, 2014].

Traditional approaches to this task normally design effective feature templates by making use of external resources like linguistic parser and sentiment lexicons, and then employ the traditional statistical learning algorithms for prediction [Kiritchenko *et al.*, 2014; Wagner *et al.*, 2014; Vo and Zhang, 2015]. For example, Vo and Zhang [2015] manually designed rich features including sentiment-specific word embedding and sentiment lexicons. Although these methods have achieved a comparable performance, their results highly depended on the effect of the handcraft features.

In recent years, neural network approaches are of growing attention for their capacity of encoding sentence in continuous and low-dimensional vector without feature engineering. Kinds of neural network methods, such as Recursive Neural Network [Socher *et al.*, 2011; Dong *et al.*, 2014; Qian *et al.*, 2015], convolutional neural network [Kalchbrenner *et al.*, 2014; Kim, 2014], LSTM [Hochreiter and Schmidhuber, 1997] and tree-structured LSTM [Tai *et al.*, 2015] were applied into the field of sentiment analysis and opinion mining, including aspect-based sentiment classification. However, most of these neural network based approaches just make use of the review context, but ignored the consideration of the target information which is supposed to be very important in indicating the aspect's sentiment polarity.

The state-of-the-art works in aspect-based sentiment classification pay more attention to incorporating the target information into the model. Tang et al. [2016a] proposed TD-LSTM and TC-LSTM which develop two target dependent long short-term memory to model the left and right contexts with target, where target information is automatically taken into account. Wang et al. [2016] proposed an attention-based LSTM to concentrate on different parts of a sentence when different targets are taken as input, but the final improvement is limited. Meanwhile, Tang et al. [2016b] designed a deep memory networks which consist of multiple computational layers to integrate the aspect information. Each layer is a context- and location- based attention model, which first learns the importance/weight of each context word and then utilizes the information to calculate context representation. Zhang et al. [2016] used one gated-RNN to learn the representation of sentence in three components and meanwhile use a gated mechanism to leverage the interaction of targets and contexts. Ma et al. [2017] proposed an interactive attention networks which interactively learn attentions in the contexts and targets. Different from previous models, our model use three LSTMs to model left context, target phrase and right context separately; meanwhile, considering the interaction of targets and contexts, we propose a rotatory attention mechanism.

## 3 Model

The overall architecture of the proposed LCR-Rot model is shown in Figure 1.

### 3.1 Left-Center-Right Separated LSTMs

Firstly, a sentence is separated into three parts, i.e., left context, target phrase and right context. Suppose a sentence $s$ contains $N$ words $[w_1, w_2, \ldots, w_N]$, and is separated into three parts: left context $[w_1^l, w_2^l, \ldots, w_L^l]$, target phrase $[w_1^t, w_2^t, \ldots, w_M^t]$ and right context $[w_1^r, w_2^r, \ldots, w_R^r]$, where $L$, $M$, $R$ is the length of three parts respectively. The sum of $L$, $M$, $R$ is equal to $N$. A unit of the three component parts will be considered as a training/testing example in the network.

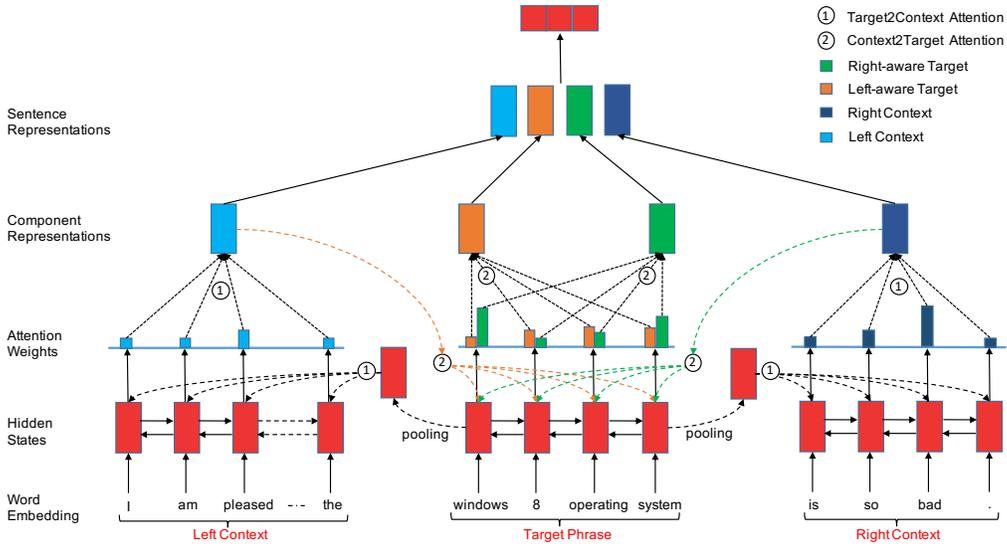

Figure 1: The Architecture of LCR-Rot. After encoding three parts of the sentence by three Bi-LSTMs, ① use target2context attention to obtain the representations of left and right context; ② then use context2target attention to obtain left-aware target and right-aware target. The attention weights in left-aware and right-aware targets are different. A combination of the two are called a two-side representation of target. Target2context and context2target constitute our rotary attention. The sentence representation consists of four component representations. The attention weights shown in the figure are the ideal values we expect.

Taking Example 1 for instance, with respect to the target "*the life of battery*", left context is "*i am pleased with*", target phrase is "*the life of battery*", right context is "*, but the windows 8 operating system is so bad.*"; with respect to the target "*windows 8 operating system*", left context is "*i am pleased with the life of battery, but the*", target phrase is "*windows 8 operating system*", right context is "*is so bad.*".

Accordingly, the LCR-Rot model is composed of three Bi-LSTMs, i.e., left-, center-, and right- Bi-LSTM, respectively modeling left context, target phrase and right context in the sentence. Specifically, each word is represented as word embedding [Bengio *et al.*, 2003; Mikolov *et al.*, 2013]. All the word vectors are stacked in a word embedding matrix $L_w \in R^{d \times |V|}$, where $d$ is the dimension of word vector and $|V|$ is vocabulary size. After we feed word embedding to Bi-LSTM, we can get hidden states $[h_1^l, h_2^l, \ldots, h_L^l]$ for left context, $[h_1^t, h_2^t, \ldots, h_M^t]$ for the target phrase and $[h_1^r, h_2^r, \ldots, h_R^r]$ for right context as the initial representations.

### 3.2 Rotatory Attention Mechanism

Secondly, a rotatory attention mechanism is designed to capture the most indicative words in target and left/right contexts. The rotatory attention mechanism contains two steps. First step, the target2context attention is used to capture the most indicative sentiment words in left/right contexts; second step, based on the new representations of left/right contexts, the context2target attention is further constructed to capture the most important word in the target and finally get a two-side representation of the target.

**(1) Target2Context Attention**

We first make use of an average representation of the target to obtain better representations of left and right contexts. An average pooling operation is used to obtain the simple representation of target phrase:

$$r^t = \text{pooling}([h_1^t, h_2^t, \ldots, h_M^t]). \quad (1)$$

In order to obtain the representation of the left and right components respectively, we first define a score function $f$ by using the hidden states of each word in the context $h_i^l(h_i^r)$ and the average pooling of the target phrase $r^t$ as inputs (taking left context for example):

$$f(h_i^l, r^t) = \tanh(h_i^l \cdot W_c^l \cdot r^t + b_c^l), \quad (2)$$

where $W_c^l$ and $b_c^l$ are weight matrix and bias respectively, and *tanh* is a non-linear function.

The score $f$ is used as a weight that denotes the importance of a word in the context indicating the sentiment toward a target. On this basis, the normalized importance weight $\alpha_i$ in the left contexts are computed as follows:

$$\alpha_i^l = \frac{\exp(f(h_i^l, r^t))}{\sum_{j=1}^{L} \exp(f(h_j^l, r^t))}. \quad (3)$$

At last, a weighted combination of word hidden states is considered as the component representation for left contexts:

$$r^l = \sum_{i=1}^{L} \alpha_i^l \cdot h_i^l. \quad (4)$$

The same as Equation (2)-(4), we can obtain $r^r$ for right context.

**(2) Context2Target Attention**

We further make use of the new representations of left/right contexts ($r^l/r^r$), to construct a better representation of the target. Just like target2context attention, we first define a score function $f$ by using the hidden states of each word in the target phrase $h_i^t$ and the final representations $r^l/r^r$ of left/right context as inputs (taking left context for example):

$$f(h_i^t, r^l) = \tanh(h_i^t \cdot W_t^l \cdot r^l + b_t^l), \quad (5)$$

where $W_t^l$ and $b_t^l$ are weight matrix and bias respectively, and *tanh* is a non-linear function.

The score $f$ is used as a weight that denotes the importance of a word in the target phrase influenced by left context. On this basis, the normalized importance weight $\alpha_i$ are computed as follows:

$$\alpha_i^{t_l} = \frac{\exp(f(h_i^t, r^l))}{\sum_{j=1}^{M} \exp(f(h_j^t, r^l))}. \quad (6)$$

At last, a weighted combination of target phrase hidden states are computed:

$$r^{t_l} = \sum_{i=1}^{M} \alpha_i^{t_l} \cdot h_i^t, \quad (7)$$

which we call left-aware target representation. Similar as Equation (5)-(7), we can get the right-aware target representation, $r^{t_r}$.

We name the combination of left-aware and right-aware target representations $[r^{t_l}, r^{t_r}]$ as the two-side representation of the target.

### 3.3 Representation Concatenation

Finally, we concatenate the left-context representation $r^l$, right-context representation $r^r$, and the two-side target representation $[r^{t_l}, r^{t_r}]$, and use it as the final representation for the sentence:

$$v = [r^l; r^{t_l}; r^{t_r}; r^r]. \quad (8)$$

The sentence representation $v$ is feed to a softmax function for aspect-level sentiment prediction:

$$p = \text{softmax}(W_c \cdot v + b_c), \quad (9)$$

where $W_c$ and $b_c$ are the parameters of the softmax layer.

### 3.4 Model Training

The model is trained in a supervised manner by minimizing the cross entropy error of sentiment classification. The loss function with respect to one training instance is defined as:

$$L = -\sum_{i=1}^{C} y_i \log(p_i) + \lambda \parallel \Theta \parallel^2, \quad (10)$$

where $C$ is the number of class labels; $y_i$ is one-hot class labels for the $i$-th class; $p_i$ is the predicted probability for the $i$-th class; $\lambda$ is weight of $L_2$−regularization; $\Theta$ is the parameter set which contains $\{W_c^l, b_c^l, W_c^r, b_c^r, W_t^l, b_t^l, W_t^r, b_t^r, W_c, b_c\}$

| Dataset | Pos.(#) | Neu.(#) | Neg.(#) |
|---|---|---|---|
| Restaurant-Train | 2164 | 637 | 807 |
| Restaurant-Test | 728 | 196 | 196 |
| Laptop-Train | 994 | 464 | 870 |
| Laptop-Test | 341 | 169 | 128 |
| Twitter-Train | 1561 | 3127 | 1560 |
| Twitter-Test | 173 | 346 | 173 |

Table 1: The statistics (number of examples in each class) of the three datasets.

and parameters in LSTM. But the initial word embedding vectors are not trained.

We use back propagation and stochastic gradient descent optimizer to train the model. The dropout strategy is used to avoid overfitting.

## 4 Experiments

### 4.1 Experimental Setting

We conduct experiments on three datasets, as shown in Table 1. The first two are from SemEval 2014 Task 4 [1] [Pontiki et al., 2014], one from laptop domain and another from restaurant domain. The third one is a collection of tweets, collected by [Dong et al., 2014]. The evaluation metric is classification accuracy.

In our work, the dimension of word embedding vectors and hidden state vectors is 300. We use GloVe[2] vectors with 300 dimensions to initialize the word embeddings, the same as [Wang et al., 2016; Tang et al., 2016b]. All out-of-vocabulary words and weight matrices are randomly initialized by a uniform distribution U(-0.1, 0.1), and all bias are set to zero. TensorFlow is used for implementing our neural network model. In model training, the learning rate is set to 0.1, the weight for $L_2$-norm regularization is set to 1e-5, and dropout rate is set to 0.5. We train the model use stochastic gradient descent optimizer with momentum of 0.9. The paired $t$-test is used for the significance testing.

### 4.2 Compared Systems

We compare our LCR-Rot model with the following systems:

1. **Majority** assigns the sentiment polarity that has the largest probability in the training set;
2. **Simple SVM** is a SVM classifier with simple features such as unigrams and bigrams;
3. **Feature-enhanced SVM** is a SVM classifier with a state-of-the-art feature template which contains n-gram features, parse features and lexicon features [Kiritchenko et al., 2014];
4. **LSTM** represents a standard LSTM for aspect-based sentiment classification task [Tang et al., 2016a];
5. **TD-LSTM** adopts two LSTMs to model the left context with target and the right context with target respectively [Tang et al., 2016a];

---
[1] The introduction of SemEval 2014 Task 4 can be obtained at http://alt.qcri.org/semeval2014/task4/
[2] Pre-trained word vectors of GloVe can be downloaded at https://nlp.stanford.edu/projects/glove/

|  | **Restaurant** (%) | **Laptop** (%) | **Twitter** (%) |
|---|---|---|---|
| Majority | 53.50 | 65.00 | 50.00 |
| Simple SVM | 73.22 | 66.97 | 62.70 |
| Feature-enhanced SVM | 80.90 | 72.10 | 71.10 |
| LSTM (Tang, 2016a) | 74.30 | 66.50 | 66.50 |
| TD-LSTM (Tang, 2016a) | 75.60 | 68.10 | 70.80 |
| AE-LSTM (Wang, 2016) | 76.60 | 68.90 | - |
| ATAE-LSTM (Wang, 2016) | 77.20 | 68.70 | - |
| GRNN-G3(Zhang, 2016) | 79.55* | 71.47* | 70.09* |
| MemNet (Tang, 2016b) | 79.98* | 70.33* | 70.52* |
| IAN (Ma, 2017) | 78.60 | 72.10 | - |
| LCR-Rot (our approach) | **81.34** | **75.24** | **72.69** |

Table 2: The performance (classification accuracy) of different methods on three datasets. The results with * are obtained by running the code posted at original paper.

|  | **Restaurant** (%) | **Laptopt** (%) | **Twitter** (%) |
|---|---|---|---|
| No-Target-Attention | 81.07 | 74.45 | 72.11 |
| No-Target-Learned | 78.04 | 70.06 | 68.21 |
| LCR-Rot | **81.34** | **75.24** | **72.69** |

Table 3: The performance of LCR-Rot and two target-reduced versions of LCR-Rot.

|  | **Single-word** (len=1) | **Multi-word** (len=2) | **Multi-word** (len>2) |
|---|---|---|---|
| Restaurant | 3521/74.5% | 819/17.3% | 388/8.2% |
| Laptop | 1825/61.5% | 857/28.9% | 284/9.6% |
| Twitter | 2081/30.0% | 4852/69.9% | 7/0.1% |

Table 4: The number/percentage of single-word and multi-word targets on the datasets.

6. **AE-LSTM** is an upgraded version of **LSTM**. For each word in a sentence, this model appends target embedding to it. Then feed these embeddings to LSTM [Wang *et al.*, 2016];

7. **ATAE-LSTM** is developed based on **AE-LSTM**. It further strengthens the effect of target embedding, which appends target embedding to each word hidden vector and leverages attention mechanism to obtain weights of each hidden vector [Wang *et al.*, 2016];

8. **GRNN-G3** adopts a Gated-RNN to represent sentence and use a three-way structure to leverage contexts [Zhang *et al.*, 2016].

9. **MemNet** is a deep memory network which considers the content and position of target [Tang *et al.*, 2016b].

10. **IAN** interactively learns attentions in the contexts and targets, and generate the representations for targets and contexts separately [Ma *et al.*, 2017].

### 4.3 System Performance Comparison

The performance of all compared systems are reported in Table 2.

We can find that the Majority method is the worst, which means the majority sentiment polarity occupies 53.50%, 65.00% and 50% of all samples on the Restaurant, Laptop and Twitter testing datasets respectively. The Simple SVM model performs better than Majority. With the help of feature engineering, the Feature-enhanced SVM achieves much better results. However, feature engineering is labor-intensive and almost reaches its performance bottleneck. Our model achieves significantly better results than feature-enhanced SVM. It shows that neural networks can obtain better representations of sentence without manual feature engineering.

Among LSTM based neural networks described in this paper, the basic LSTM approach performs the worst. TD-LSTM obtains an improvement of 1-2% over LSTM when target signals are taken into consideration. Because of the introduction of attention mechanism, AE-LSTM and ATAE-LSTM achieve better results than TD-LSTM, and ATAE-LSTM is slightly better among the two. IAN, GRNN-G3 and MemNet show different advantages in different datasets. MemNet achieves better results than other models on the Restaurant dataset, since it considers not only the contexts of targets but also the position of each context word related to the target. IAN considers separate representations of targets and obtains better result on the Laptop dataset. GRNN-G3 achieves competitive results on all datasets because of its three-way structure and special gated-RNN model.

In the contrast, our LCR-Rot model achieves the best results on the all datasets among all models. And we will give a detailed analysis in the following subsections.

### 4.4 The Effect of Two-side Target Representation

In order to verify the effectiveness and advantage of our two-side target representation, we design the following two reduced models based on **LCR-Rot**:

1. **No-Target-Attention** is a simplified version of **LCR-Rot**, where we remove context2target attention and use the average of hidden states of target phrase to represent the target phrase;

2. **No-Target-Learned** is based on **No-Target-Attention**, where the target phrase is not learned by a LSTM independently. Instead, the average of initial word embeddings is used to represent the target phrase.

In Table 3, we report the performance of LCR-Rot and two target-reduced models. It can be seen that **No-Target-Attention** model performs a little worse than **LCR-Rot**. The results verify the usage of target attention is rewarding in our model. By comparing **LCR-Rot** and **No-Target-Learned**, we find that removing the support of multi-word target phrase will cause a more rapid decline. The decreases are 3.30%, 5.18% and 4.48% on the Restaurant, Laptop and Twitter dataset respectively. It indicates that the two-side way to model the target phrase is very important for aspect-based sentiment classification.

To analyze the problem more deeply, we summarize the number/percentage of single-word targets and multi-word targets on the datasets in Table 4. It can be seen that more than 1/4 of targets on the Restaurant contain multiple words. The percentage is even more than 1/3 and 2/3 on the Laptop and Twitter respectively. It is reasonable that the decreases of **No-

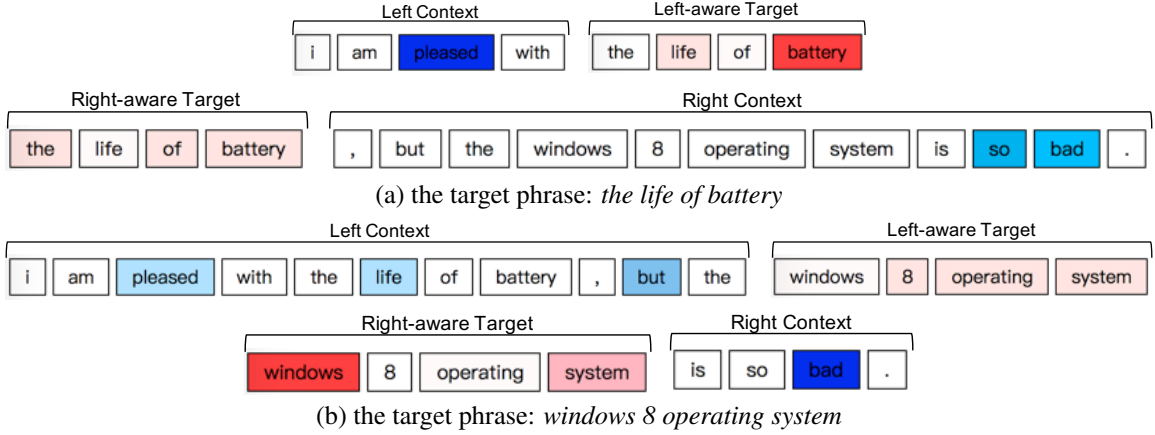

(a) the target phrase: *the life of battery*

(b) the target phrase: *windows 8 operating system*

Figure 2: Attention visualizations of Example 1, with the targets of "*the life of battery*" and "*windows 8 operating system*" respectively.

|  | Restaurant (%) | Laptopt (%) | Twitter (%) |
| --- | --- | --- | --- |
| No-Attention | 79.02 | 71.79 | 70.66 |
| Attention-Reverse | 79.73 | 74.45 | 72.11 |
| LCR-Rot | **81.34** | **75.24** | **72.69** |

Table 5: The performance of LCR-Rot and attention changed versions of LCR-Rot.

**Target-Learned** on the Laptop and Twitter are more than that on the Restaurant.

In summary, our two-side target representation way is effective to support multi-word target representation.

### 4.5 The Effect of Rotatory Attention

To verify the effectiveness of rotatory attention, we further design the following models based on **LCR-Rot**:

1. **No-Attention** is based on **No-Target-Attention**. We continue to remove the target2context attention mechanism in **No-Target-Attention** and use the average of hidden states to represent left and right contexts;

2. **Attention-Reverse** is based on **LCR-Rot**, where we reverse the order of attention. We first adopt context2target attention and then adopt target2context attention.

We have known that context2target attention is rewarding in our model according to previous subsection. When the attention is further reduced, from Table 5, we can see that the performance of **No-Attention** drops significantly. It illustrates target2context attention is more important than context2target attention. By comparing **LCR-Rot** and **Attention-Reverse**, we find that reversing the order of attention will cause 0.5%-1.5% performance decrease on the datasets which proves the advantage of our rotatory attention.

In order to acquire a better understanding of the left-center-right separated rotatory attention model, we propose a visualization toolkit to show the attention weights (Equation 3 and 6) of contexts and target phrases. In Figure 2, we give the visualization of Example 1. The red color denotes words in the target phrases and the blue color denotes words in the contexts to which the model pays attention. The darker of the color, the more important of the word for the representation. We observe Figure 2 from several angles.

Firstly, seen from both sub-figures, the most indicative sentiment word in the context can be accurately captured. For example, in sub-figure (a), the word "*pleased*" has the biggest attention weight for the target "*the life of battery*". Meanwhile, in sub-figure (b), given the target "*windows 8 operation system*", although both left and right contexts contain sentiment word ("*pleased*" in the left context and "*bad*" in the right context), the correct sentiment word "*bad*" in the right context is selected as the most important one.

Secondly, in sub-figure (a) and (b), we can see that attention weights of left-aware target phrase and right-aware target phrase are very different. When the target phrase is more related to left contexts, the attention weights of left-aware target phrase is more suitable. This may be a special explanation of the effectiveness of rotatory attention.

Thirdly, by comparing ideal values of attention weights that we expect in Figure 1 and real values that we shown in Figure 2 with respect to the target "*windows 8 operating system*", we can see that real values and ideal values are generally the same. This shows that our model behaves as we expected.

## 5 Conclusions

In this paper, we propose a left-center-right separated neural network with rotatory attention model for aspect-based sentiment analysis. The key idea of our model is to represent a sentence with a specific target as the concatenation of left-center-right component representations. Under such a network framework, we further propose a rotatory attention mechanism to take into account the interaction between targets and contexts to better represent targets and contexts. The experimental results on three benchmark datasets demonstrate that our model achieves currently the best aspect-based sentiment classification performance, in comparison with the state-of-the-art methods proposed in recent years.

# References

[Bengio *et al.*, 2003] Yoshua Bengio, Réjean Ducharme, Pascal Vincent, and Christian Jauvin. A neural probabilistic language model. *Journal of machine learning research*, 3(Feb):1137–1155, 2003.

[Dong *et al.*, 2014] Li Dong, Furu Wei, Chuanqi Tan, Duyu Tang, Ming Zhou, and Ke Xu. Adaptive recursive neural network for target-dependent twitter sentiment classification. In *Proceedings of the 52nd Annual Meeting of the Association for Computational Linguistics*, pages 49–54, 2014.

[Hochreiter and Schmidhuber, 1997] Sepp Hochreiter and Jürgen Schmidhuber. Long short-term memory. *Neural computation*, 9(8):1735–1780, 1997.

[Jiang *et al.*, 2011] Long Jiang, Mo Yu, Ming Zhou, Xiaohua Liu, and Tiejun Zhao. Target-dependent twitter sentiment classification. In *Proceedings of the 49th Annual Meeting of the Association for Computational Linguistics: Human Language Technologies-Volume 1*, pages 151–160. Association for Computational Linguistics, 2011.

[Kalchbrenner *et al.*, 2014] Nal Kalchbrenner, Edward Grefenstette, and Phil Blunsom. A convolutional neural network for modelling sentences. In *ACL*, pages 655–665, 2014.

[Kim, 2014] Yoon Kim. Convolutional neural networks for sentence classification. In *Proceeding of the conference on empirical methods in natural language processing*, pages 1746–1751, 2014.

[Kiritchenko *et al.*, 2014] Svetlana Kiritchenko, Xiaodan Zhu, Colin Cherry, and Saif Mohammad. Nrc-canada-2014: Detecting aspects and sentiment in customer reviews. In *Proceedings of the 8th International Workshop on Semantic Evaluation (SemEval 2014)*, pages 437–442, 2014.

[Liu, 2012] Bing Liu. Sentiment analysis and opinion mining. *Synthesis lectures on human language technologies*, 5(1):1–167, 2012.

[Ma *et al.*, 2017] Dehong Ma, Sujian Li, Xiaodong Zhang, and Houfeng Wang. Interactive attention networks for aspect-level sentiment classification. In *IJCAI*, 2017.

[Mikolov *et al.*, 2013] Tomas Mikolov, Ilya Sutskever, Kai Chen, Greg S Corrado, and Jeff Dean. Distributed representations of words and phrases and their compositionality. In *Advances in neural information processing systems*, pages 3111–3119, 2013.

[Pang and Lee, 2008] Bo Pang and Lillian Lee. Opinion mining and sentiment analysis. *Foundations and Trends® in Information Retrieval*, 2(1–2):1–135, 2008.

[Pontiki *et al.*, 2014] Maria Pontiki, Dimitris Galanis, John Pavlopoulos, Harris Papageorgiou, Ion Androutsopoulos, and Suresh Manandhar. Semeval-2014 task 4: Aspect based sentiment analysis. *Proceedings of SemEval*, pages 27–35, 2014.

[Qian *et al.*, 2015] Qiao Qian, Bo Tian, Minlie Huang, Yang Liu, Xuan Zhu, and Xiaoyan Zhu. Learning tag embeddings and tag-specific composition functions in recursive neural network. In *Proceedings of the 53th Annual Meeting of the Association for Computational Linguistics*, pages 1365–1374, 2015.

[Socher *et al.*, 2011] Richard Socher, Jeffrey Pennington, Eric H Huang, Andrew Y Ng, and Christopher D Manning. Semi-supervised recursive autoencoders for predicting sentiment distributions. In *Proceedings of the conference on empirical methods in natural language processing*, pages 151–161. Association for Computational Linguistics, 2011.

[Tai *et al.*, 2015] Kai Sheng Tai, Richard Socher, and Christopher D Manning. Improved semantic representations from tree-structured long short-term memory networks. In *ACL*, 2015.

[Tang *et al.*, 2016a] Duyu Tang, Bing Qin, Xiaocheng Feng, and Ting Liu. Effective lstms for target-dependent sentiment classification. In *International Conference on Computational Linguistics*, pages 3298–3307, 2016.

[Tang *et al.*, 2016b] Duyu Tang, Bing Qin, and Ting Liu. Aspect level sentiment classification with deep memory network. In *Proceeding of the conference on empirical methods in natural language processing*, pages 214–224, 2016.

[Vo and Zhang, 2015] Duy-Tin Vo and Yue Zhang. Target-dependent twitter sentiment classification with rich automatic features. In *IJCAI*, pages 1347–1353, 2015.

[Wagner *et al.*, 2014] Joachim Wagner, Piyush Arora, Santiago Cortes, Utsab Barman, Dasha Bogdanova, Jennifer Foster, and Lamia Tounsi. Dcu: Aspect-based polarity classification for semeval task 4. In *Proceedings of the 8th International Workshop on Semantic Evaluation (SemEval 2014)*, pages 223–229, 2014.

[Wang *et al.*, 2016] Yequan Wang, Minlie Huang, Li Zhao, and Xiaoyan Zhu. Attention-based lstm for aspect-level sentiment classification. In *Proceeding of the conference on empirical methods in natural language processing*, pages 606–615. Association for Computational Linguistics, 2016.

[Zhang *et al.*, 2016] Meishan Zhang, Yue Zhang, and Duy-Tin Vo. Gated neural networks for targeted sentiment analysis. In *AAAI*, pages 3087–3093, 2016.